\documentclass{article}
\usepackage[utf8]{inputenc} 
\usepackage[T1]{fontenc}    
\usepackage[pagebackref, breaklinks, colorlinks]{hyperref}       
\usepackage{url}            
\usepackage{booktabs}       
\usepackage{amsfonts}       
\usepackage{nicefrac}       
\usepackage{microtype}      
\usepackage{xcolor}         
\usepackage{graphicx}
\usepackage{multirow}
\usepackage{colortbl}
\usepackage{subcaption}
\usepackage{amsmath}
\usepackage{cleveref}
\usepackage{wrapfig}
\usepackage{float}

\title{PanguMotion: Continuous Driving Motion Forecasting with Pangu Transformers}

\author{
  Quanhao Ren$^{*}$
  \quad 
  \quad 
  Yicheng Li\thanks{Equal contribution.}
  \quad 
  \quad
  Nan Song
  \vspace{.4em} 
  \\
  $^{1}$School of Data Science, Fudan University
  \quad 
  \\
  \url{https://github.com/QuanhaoR/RealMotionPanGu}
}

\begin{document}

\maketitle


\begin{abstract}

\textbf{Motion forecasting} is a core task in autonomous driving systems, aiming to accurately predict the future trajectories of surrounding agents to ensure driving safety. Existing methods typically process discrete driving scenes independently, neglecting the temporal continuity and historical context correlations inherent in real-world driving environments. This paper proposes \textbf{PanguMotion}, a motion forecasting framework for continuous driving scenarios that integrates \textbf{Transformer blocks from the Pangu-1B large language model} as feature enhancement modules into autonomous driving motion prediction architectures. We conduct experiments on the \textit{Argoverse 2 datasets processed by the RealMotion data reorganization strategy}, transforming each independent scene into a continuous sequence to mimic real-world driving scenarios.

\textbf{PanguMotion} introduces two key innovations based on the RealMotion framework: First, we insert a frozen Pangu Transformer block after the encoder as a feature enhancer, leveraging the powerful representation capabilities of pre-trained large models while removing the agent trajectory stream to avoid functional redundancy. Second, we implement a complete \textbf{Ascend NPU} adaptation and training solution, providing an end-to-end inference implementation optimized for NPU architecture.

Experimental results demonstrate that our approach achieves significant improvements in continuous driving motion forecasting, particularly in single-mode prediction metrics, validating the generalization capability of large language models in non-linguistic sequential tasks. We also open-source the complete implementation code, providing a new research baseline and technical reference for the autonomous driving field. Code is available on GitHub.
\end{abstract}

\section{Introduction}
\label{sec:intro}

Motion forecasting is critical for autonomous driving safety\cite{wilson2023argoverse}. Most existing methods treat driving scenes as independent units\cite{phan2020covernet}, overlooking the continuous nature of real-world driving. Song et al.\cite{song2024motion} proposed RealMotion, the first framework to model continuous scenes through data reorganization and dual-stream design, yet the potential of pre-trained large models for enhancing scene representation in continuous driving scenarios remains under-explored.

Concurrently, Pang et al.\cite{pang2024frozen} discovered that LLM Transformer blocks can directly serve as visual encoders, demonstrating excellent "information filtering" capabilities across various vision tasks. This inspires our research: Can LLMs be integrated into continuous driving forecasting frameworks to improve motion prediction performance?

This paper proposes \textbf{PanguMotion}, the first framework to integrate a Pangu-1B\cite{chen2025pangu} Transformer block into continuous driving motion prediction. As shown in Figure\ref{fig:framework}, we insert a frozen Pangu block after the encoder as a feature enhancement module within a simplified RealMotion architecture that preserves only the scene context stream.

Our main contributions are threefold:
1. \textbf{Methodological Innovation}: We are the first to successfully integrate Pangu large models into continuous driving prediction frameworks, demonstrating their effectiveness as feature enhancers for temporal prediction tasks.
2. \textbf{Architectural Optimization}: We propose a simplified architecture that removes the agent trajectory stream while maintaining scene context modeling, achieving better single-mode prediction performance through the synergistic combination of frozen LLM blocks and continuous scene representation.
3. \textbf{Practical Feasibility}: We provide a complete Ascend NPU adaptation and training solution, implementing an end-to-end pipeline from model training to inference deployment that maintains comparable latency while enhancing feature representation.

We validate our method on the Argoverse 2 datasets\cite{wilson2023argoverse} processed by RealMotion's data reorganization strategy. Experimental results demonstrate that PanguMotion achieves significant improvements in single-mode prediction metrics (minFDE$_1$1.58\% and minADE$_1$1.29\%), confirming the transferability of large language models to non-linguistic sequential tasks and their practical value in autonomous driving applications. We have open-sourced the complete implementation code to provide reproducible benchmarks for related research.

\section{Related work}
\label{sec:related_work}

\subsection{Motion Forecasting in Autonomous Driving}
Accurately predicting the future trajectories of surrounding agents is fundamental to the safety of autonomous driving. Early approaches\cite{phan2020covernet,gilles2021home,chai2020multipath} primarily relied on rasterizing driving scenarios into images and employing Convolutional Neural Networks for context encoding. However, these methods often struggled to capture intricate structural information. Consequently, the field has shifted towards\cite{zhao2021tnt,gu2021densetnt,9878832,varadarajan2022multipath++} vectorized representations, as exemplified by VectorNet, which encodes map elements and agent dynamics as polylines. Building on this, recent methods have widely adopted graph-based structures and Transformer architectures to model complex interactions and dynamics within the scene\cite{liang2020learning,gilles2022gohome,zeng2021lanercnn,jia2023hdgt,pmlr-v164-jia22a,deo2022multimodal}.

Despite these advancements, a significant limitation in most existing methods, including state-of-the-art approaches like Wayformer\cite{nayakanti2023wayformer} and Scene Transformer\cite{ngiamscene}, is that they process driving scenes as independent, isolated snapshots. This "independent" processing ignores the temporal interconnectedness inherent in real-world driving. To address this, recent pioneering work has introduced frameworks for continuous driving. Notably, the RealMotion\cite{song2024motion} framework proposes modeling continuous scenes through data reorganization and dual-stream designs, allowing the system to effectively utilize historical context from successive scenes. This builds upon the continuous forecasting paradigm, aiming to further enhance the representation of these continuous scenes.

\subsection{Large Language Models for Visual Encoding}
The evolution of Large Language Models, driven by scaling laws\cite{kaplan2020scaling} and masked token prediction, has produced models with billions of parameters, such as GPT\cite{brown2020language}, LLaMA\cite{touvron2023llama}, and Pangu. These models exhibit remarkable zero-shot performance and in-context learning capabilities. Traditionally, in the context of vision tasks, LLMs have been utilized either as text encoders within Vision-Language Models\cite{koh2023grounding,lin2024awq,merullo2024language,schwettmann2023find} or as decoders that translate latent visual features into tokenized outputs\cite{alayrac2022flamingo}. These frameworks typically rely on projecting visual features into the LLM's input space or using latent bottlenecks.

However, a more recent and distinct line of research has emerged regarding the internal capabilities of these models. Investigations have discovered that the transformer blocks within pre-trained LLMs can act as powerful, general-purpose visual encoder layers without requiring language prompts or text alignment. For instance, studies\cite{pang2024frozen} have shown that inserting a frozen transformer block directly into visual encoders can significantly enhance performance across various tasks, including image classification and motion forecasting. This phenomenon is attributed to an "information filtering" hypothesis, wherein the pre-trained LLM block identifies and amplifies informative visual tokens.

\subsection{Synergizing Continuous Forecasting with Pangu}
While RealMotion addresses the temporal continuity of driving scenes and recent works have validated the visual encoding capabilities of LLMs, the potential of leveraging the Pangu large language model within a continuous forecasting framework remains underexplored. In this work, we bridge these two domains. We adopt the continuous scene modeling of RealMotion and integrate a frozen transformer block from the Pangu-1B model. Unlike previous approaches that utilize LLMs solely for linguistic tasks or rely on LLaMA-based visual encoding, we explore Pangu's specific capability to serve as a feature enhancer for the spatiotemporal sequences inherent in autonomous driving. By leveraging Pangu's pre-trained representations, we aim to improve the feature extraction of continuous driving streams, thereby enhancing the accuracy of trajectory predictions in complex, evolving scenarios.

\section{Methodology}

\subsection{Overall Architecture}

\begin{figure}[tb]
    \centering
    \includegraphics[width=\linewidth]{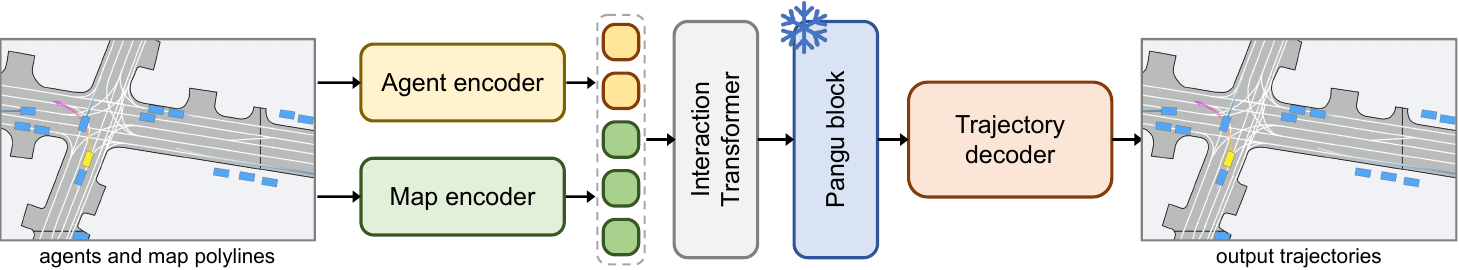} 
    \caption{Overview of our PanguMotion architecture. PanguMotion adopts an encoder-decoder structure with a frozen Pangu Block designed to capture interactive relationships within each scene and across the continuous scenes.}
    \label{fig:framework}
\end{figure}

As illustrated in Figure \ref{fig:framework}, PanguMotion builds upon the RealMotion framework, preserving the scene context stream from its continuous prediction dual-stream design but removing the agent trajectory stream. Our core innovation lies in inserting a frozen Pangu-1B Transformer block as a feature enhancement module after the self-attention encoding layers. This module is positioned after scene interaction and multiple Transformer block encoding but before decoder prediction, forming a hierarchical feature enhancement mechanism.

The complete processing pipeline can be formalized into the following steps: first, the base encoder transforms input data $X_{\text{input}}$ into initial scene features; second, the scene context stream fuses historical scene information through cross-attention mechanisms, forming preliminary scene representations; third, multiple Transformer blocks further encode scene features to obtain $F_s$; fourth, the Pangu block performs deep feature enhancement on $F_s$ to obtain $F_{\text{enhanced}}$; finally, the decoder directly outputs predicted trajectories $\hat{Y}$ based on the enhanced features. Mathematically, this is expressed as:

\begin{equation}
\begin{aligned}
F_0 &= \text{BaseEncoder}(X_{\text{in}}), \\
F_c &= \text{SceneContextStream}(F_0), \\
F_s &= \text{TransformerBlocks}(F_c), \\
F_e &= \text{PanguBlock}(F_s), \\
\hat{Y} &= \text{Decoder}(F_e).
\end{aligned}
\end{equation}

\subsection{Input Representation and Base Encoder}

We fully adhere to RealMotion's data preprocessing and input representation methodology. Specifically, we employ the data reorganization strategy proposed by RealMotion to transform independent scenes in the Argoverse datasets into continuous sequences, simulating real-world continuous driving scenarios. For each timestep, input data comprises historical trajectories $A \in \mathbb{R}^{N_a \times T \times C_a}$ and map elements $M \in \mathbb{R}^{N_m \times P \times C_m}$, containing agents' motion states and road network geometry, respectively.

The base encoder follows RealMotion's encoder design, consisting of a PointNet-like map feature encoder and an agent feature encoder based on neighborhood attention mechanisms. The outputs of these two encoders are concatenated and augmented with positional encoding and type encoding, forming initial scene feature representations $F_{\text{initial}} \in \mathbb{R}^{(N_a+N_m) \times D}$, where $D=128$ is our feature dimension.

\subsection{Scene Interaction and Encoding}

After obtaining initial scene features, the model first performs scene interaction processing. This step corresponds to RealMotion's scene context stream, fusing current scene features with historical memory features through cross-attention mechanisms. Specifically, for timestep $t$, the model aligns scene features from timestep $t-1$ to the current coordinate system, then performs scene interaction separately for agent and map features:

\begin{equation}
F_{\text{agent}}' = \text{SceneInteraction}(F_{\text{agent}}, F_{\text{memory}}, \text{pose\_info})
\end{equation}
\begin{equation}
F_{\text{map}}' = \text{SceneInteraction}(F_{\text{map}}, F_{\text{memory\_map}}, \text{pose\_info})
\end{equation}

where $F_{\text{agent}}$ and $F_{\text{map}}$ are agent and map features extracted from $F_{\text{initial}}$, $F_{\text{memory}}$ and $F_{\text{memory\_map}}$ are corresponding features from historical memory, and pose information includes time difference, angle difference, and position difference.

The interacted features are recombined into $F_{\text{scene}}$, which is then further encoded through multiple Transformer blocks. Each Transformer block uses valid token masks (key\_valid\_mask) to control the scope of attention computation, ensuring only valid scene elements participate in calculations. After processing by multiple Transformer blocks, we obtain encoded scene features $F_s$.

\subsection{Pangu Transformer Module}

The proposed Pangu Transformer module represents the core innovation of our framework. We extract the last Transformer block (layer 25) from the pre-trained Pangu-1B model as a feature enhancement module. Since the hidden dimension of the Pangu model (1536) does not match our scene feature dimension (128), we add trainable linear projection layers before and after the module for dimension alignment.

The processing pipeline of this module is as follows: first, the feature dimension is expanded from $D=128$ to $D_{\text{pangu}}=1536$ through the linear projection layer $W_{\text{before}} \in \mathbb{R}^{D \times D_{\text{pangu}}}$; next, the features are input to the Pangu Transformer layer $T_{\text{pangu}}$. This layer receives the same valid token masks (key\_valid\_mask) as the preceding Transformer blocks, ensuring consistency in attention computation. The module output is projected back to the original dimension through the linear projection layer $W_{\text{after}} \in \mathbb{R}^{D_{\text{pangu}} \times D}$, and finally normalized using layer normalization:

\begin{equation}
F_{\text{enhanced}} = \text{LayerNorm}(W_{\text{after}} \cdot T_{\text{pangu}}(W_{\text{before}} \cdot F_s))
\end{equation}

To adapt the Pangu Transformer to motion forecasting tasks, we made one important adjustment regarding the attention mechanism. Since the original Pangu design uses autoregressive masks for text generation while all tokens are available simultaneously in motion forecasting, we remove autoregressive masks and maintain the existing valid token mask mechanism from the model. Regarding positional encoding, although the Pangu model employs rotary positional encoding, we follow RealMotion's design philosophy by removing internal positional encoding within the Pangu block, relying instead on positional information already present in the base encoder.

During training, all parameters of the Pangu Transformer module remain frozen, with only the preceding and following linear projection layer parameters being trained. This design leverages the powerful representational capabilities of pre-trained large models while avoiding overfitting risks associated with fine-tuning large model parameters.

\subsection{Decoder Design}

The features enhanced by the Pangu module $F_{\text{enhanced}}$ are directly fed into the decoder. Unlike the complete RealMotion framework, our design removes the agent trajectory stream, as experiments revealed overlapping functionality between the feature enhancement capability of the Pangu block and the refinement function of the agent trajectory stream.The decoder comprises three main components: a multimodal projection layer, a trajectory regression branch, and a probability prediction branch. The multimodal projection layer expands input features into representations for 6 modes; the trajectory regression branch is a three-layer perceptron predicting future trajectory coordinates for each mode; similarly, the probability prediction branch is a three-layer perceptron estimating confidence scores for each mode.

The loss function follows RealMotion's design, comprising regression loss $\mathcal{L}_{\text{reg}}$ and classification loss $\mathcal{L}_{\text{cls}}$, computed using smooth L1 loss and cross-entropy loss, respectively:

\begin{equation}
\mathcal{L} = \mathcal{L}_{\text{reg}} + \mathcal{L}_{\text{cls}}
\end{equation}

\subsection{Ascend NPU Adaptation}

To achieve efficient training and inference on Ascend NPU, we completed a series of adaptation tasks. For the training framework, based on the PyTorch Lightning framework, we set the accelerator to Ascend NPU and adopted a distributed data parallel strategy to fully utilize the parallel computing capabilities of multiple NPUs. Regarding software compatibility, to ensure stable framework operation, we replaced relevant software packages in the standard environment with Ascend NPU-specific versions.

Since the built-in attention mechanisms in PyTorch and the neighborhood attention mechanism (NeighborhoodAttention1D)\cite{hassani2023neighborhood} from natten require GPU support, we replaced them with standard attention mechanisms compatible with NPU. This modification ensures stable operation and efficient execution of the model on the Ascend platform.

Our implementation maintains inference latency comparable to the original RealMotion while enhancing feature representation through the Pangu module, providing an efficient NPU solution for continuous driving motion forecasting.

\section{Experiments}

\subsection{Experimental Setup}

\paragraph{Datasets and Evaluation Metrics:} We evaluate our method on the Argoverse 2 motion forecasting dataset. Argoverse 2 contains 250,000 scenes spanning six cities, offering improved data diversity and higher quality compared to its predecessor. The dataset provides a 5-second historical observation window and requires forecasting 6 seconds into the future, with all data sampled at 10 Hz frequency. Following RealMotion's methodology, we apply its data reorganization strategy to transform independent scenes into continuous sequences, simulating real-world continuous driving scenarios.

We employ standard motion forecasting evaluation metrics. Minimum Average Displacement Error (minADE$_k$) measures the average distance between predicted and ground truth trajectories across all timesteps, while minimum Final Displacement Error (minFDE$_k$) focuses specifically on endpoint accuracy at the final timestep. The subscript $k$ indicates the number of prediction modes considered: $k=1$ evaluates the single most probable trajectory, reflecting practical decision-making accuracy; $k=6$ considers the top six predictions, assessing coverage capability. Additional metrics include Miss Rate (MR$_k$) and Brier minimum Final Displacement Error (b-minFDE$_k$) for uncertainty calibration assessment.

\paragraph{Implementation Details:} Our model is implemented based on the PyTorch Lightning framework and trained on four Huawei Ascend 910B2 NPUs. The input feature dimension is set to 128. The base encoder uses 4 Transformer blocks with 8 attention heads. The Pangu Transformer module employs the last Transformer block (layer 25) of the pre-trained Pangu-1B model as a feature enhancement module. Training adopts the AdamW\cite{loshchilovdecoupled} optimizer with a learning rate of 0.001, batch size of 32 per GPU, and runs for 80 epochs to ensure sufficient convergence.

\begin{table}[t]
\centering
\caption{Performance Comparison on Argoverse 2 Validation Set}
\label{tab:av2_val}
\resizebox{\textwidth}{!}{
\begin{tabular}{c|ccccc}
\bottomrule[1.5pt]
Model & \rotatebox[origin=c]{0}{minADE$_1$} & \rotatebox[origin=c]{0}{minFDE$_1$} & \rotatebox[origin=c]{0}{minADE$_6$} & \rotatebox[origin=c]{0}{minFDE$_6$} & \rotatebox[origin=c]{0}{b-minFDE$_6$} \\
\midrule
RealMotion-I & 1.808 & 4.510 & 0.728 & 1.420 & 2.043 \\
RealMotion & 1.646 & 4.100 & \textbf{0.669} & \textbf{1.303} & \textbf{1.935} \\
\addlinespace[2pt]
\textbf{PanguMotion-I} & 1.777 & 4.435 & 0.717 & 1.398 & 2.033 \\
\textbf{PanguMotion} & \textbf{1.620} & \textbf{4.047} & 0.672 & 1.314 & 1.958 \\
\bottomrule[1.5pt]
\end{tabular}
}
\end{table}

\subsection{Main Results}

\paragraph{Argoverse 2 Validation Set:} As shown in Table \ref{tab:av2_val}, our method demonstrates excellent performance across multiple metrics. Compared to RealMotion, PanguMotion achieves improvements of 1.58\% and 1.29\% in minADE$_1$ and minFDE$_1$, respectively. Notably, PanguMotion-I (without data reorganization and agent trajectory stream) outperforms RealMotion-I across all metrics, validating the universal enhancement capability of Pangu Transformers.

\subsection{Ablation Studies and Analysis}

\subsubsection{Comparison of Different Large Language Models}

\begin{table}[t]
\centering
\caption{Comparison between Pangu and LLaMA Models}
\label{tab:llama_vs_pangu}
\resizebox{\textwidth}{!}{
\begin{tabular}{c|ccccc}
\bottomrule[1.5pt]
Model & \rotatebox[origin=c]{0}{minADE$_1$} & \rotatebox[origin=c]{0}{minFDE$_1$} & \rotatebox[origin=c]{0}{minADE$_6$} & \rotatebox[origin=c]{0}{minFDE$_6$} & \rotatebox[origin=c]{0}{b-minFDE$_6$} \\
\midrule
LlamaMotion & 1.718 & 4.265 & 0.690 & 1.339 & 1.983 \\
\textbf{PanguMotion} & \textbf{1.620} & \textbf{4.047} & \textbf{0.672} & \textbf{1.314} & \textbf{1.958} \\
\bottomrule[1.5pt]
\end{tabular}
}
\end{table}

We compare PanguMotion with LlamaMotion (using LLaMA model). As shown in Table \ref{tab:llama_vs_pangu}, the Pangu model performs significantly better than LLaMA in motion forecasting tasks, particularly in single-mode prediction metrics (minADE$_1$ improved by 5.71\%, minFDE$_1$ by 5.11\%). This suggests that Pangu's pre-training on large-scale Chinese corpora may contribute to better understanding of spatial relationships and temporal patterns, potentially because Chinese linguistic structures encode spatial information differently than English. The observed performance difference could also stem from other factors including model architecture variations, pre-training data characteristics, or domain-specific optimizations in Pangu compared to LLaMA.

\subsubsection{Impact of Different Pangu Layer Counts}

\begin{table}[t]
\centering
\caption{Comparison of Different Pangu Layer Counts}
\label{tab:layer_count}
\resizebox{\textwidth}{!}{
\begin{tabular}{c|ccccc}
\bottomrule[1.5pt]
Layer Selection & \rotatebox[origin=c]{0}{minADE$_1$} & \rotatebox[origin=c]{0}{minFDE$_1$} & \rotatebox[origin=c]{0}{minADE$_6$} & \rotatebox[origin=c]{0}{minFDE$_6$} & \rotatebox[origin=c]{0}{b-minFDE$_6$} \\
\midrule
Last layer (layer 25) & \textbf{1.620} & \textbf{4.047} & \textbf{0.672} & \textbf{1.314} & \textbf{1.958} \\
Last three layers (layers 23-25) & 1.693 & 4.204 & 0.694 & 1.359 & 2.003 \\
Last six layers (layers 20-25) & 1.702 & 4.217 & 0.697 & 1.357 & 2.002 \\
\bottomrule[1.5pt]
\end{tabular}
}
\end{table}

We investigate the effect of using different numbers of Pangu Transformer layers (Table \ref{tab:layer_count}). Experimental results clearly demonstrate that using a single last layer (layer 25) achieves the best performance across all evaluation metrics, validating the "information filtering hypothesis"\cite{pang2024frozen}: higher-layer Transformer blocks are more suitable for processing high-level semantic information. Interestingly, as the number of layers increases (from single layer to three layers to six layers), model performance shows a systematic declining trend. Compared to using a single layer, using the last three layers results in a 4.51\% increase in minADE$_1$ and a 3.88\% increase in minFDE$_1$; these metrics further deteriorate when using the last six layers. This suggests that while increasing the number of pre-trained Transformer layers enhances model capacity, it may introduce redundant information or inter-layer conflicts that actually interfere with the motion prediction task. This finding emphasizes that for feature enhancement applications, a carefully selected single layer is more effective than stacking multiple pre-trained layers, and indicates that higher-layer Transformer blocks can provide semantic representations most suitable for cross-task transfer.

\subsubsection{Impact of Different Insertion Positions}

\begin{table}[t]
\centering
\caption{Comparison of Different Insertion Positions}
\label{tab:insert_position}
\resizebox{\textwidth}{!}{
\begin{tabular}{c|ccccc}
\bottomrule[1.5pt]
Insertion Position & \rotatebox[origin=c]{0}{minADE$_1$} & \rotatebox[origin=c]{0}{minFDE$_1$} & \rotatebox[origin=c]{0}{minADE$_6$} & \rotatebox[origin=c]{0}{minFDE$_6$} & \rotatebox[origin=c]{0}{b-minFDE$_6$} \\
\midrule
Layer 0 & 1.650 & 4.107 & 0.676 & 1.313 & 1.955 \\
Layer 8 & 1.680 & 4.169 & 0.680 & 1.326 & 1.969 \\
Layer 16 & 1.677 & 4.182 & 0.682 & 1.329 & 1.972 \\
\textbf{Layer 25 (last)} & \textbf{1.620} & \textbf{4.047} & \textbf{0.672} & \textbf{1.314} & \textbf{1.958} \\
\bottomrule[1.5pt]
\end{tabular}
}
\end{table}

We investigate the effect of inserting the Pangu Transformer at different positions within the model (Table \ref{tab:insert_position}). The experimental results clearly show a trend where performance improves progressively as we select higher layers for insertion. Among all positions tested, inserting the last layer (layer 25) achieves the best overall performance across all metrics, with particularly notable improvements in single-mode prediction (minADE$_1$ improved by 1.82\% and minFDE$_1$ by 1.46\% compared to layer 0). This progression supports the "information filtering hypothesis" that higher Transformer layers encode more abstract, semantically-rich information that is better suited for feature enhancement in motion forecasting tasks.

Interestingly, we observe a U-shaped pattern in the performance curve: while layer 25 performs best, layer 0 (the earliest layer) shows better results than intermediate layers (8 and 16). This suggests that the earliest layer, though basic in its representations, may contain foundational structural patterns that are still useful for motion prediction, whereas intermediate layers might encode language-specific abstractions less transferable to trajectory forecasting. The consistent superiority of the final layer indicates that the most advanced semantic processing capabilities of pre-trained language models are most beneficial when applied to already-processed scene features, allowing for sophisticated refinement of the learned representations.
\subsubsection{Impact of Agent Trajectory Stream}

\begin{table}[t]
\centering
\caption{Ablation Study on Agent Trajectory Stream,``SC Strm'' and ``AT Strm'' indicate our proposed scene context stream and agent trajectory stream, respectively.``PanGu`` indicate the last transformer layer of PanGu Large Language Model.}
\label{tab:stream_ablation}
\resizebox{\textwidth}{!}{
\begin{tabular}{c|ccc|ccccc}
\bottomrule[1.5pt]
\multirow{2}{*}{ID}  & SC  &  AT &PanGu&  \multirow{2}{*}{$\textit{minADE}_{1}$}  &\multirow{2}{*}{$\textit{minFDE}_{1}$}& \multirow{2}{*}{$\textit{minADE}_{6}$} &\multirow{2}{*}{$\textit{minFDE}_{6}$}  &  \multirow{2}{*}{$\textit{b-minFDE}_{6}$}\\
         &Strm&Strm&&&&&&\\
\midrule
1 &\mbox{}&\mbox{}&\mbox{} & 1.793 & 4.499& 0.721 &  1.423 & 2.054 \\
2 &\mbox{}&\mbox{}&$\checkmark$ &1.777  & 4.435& 0.717 &  1.398 & 2.033 \\
3 &$\checkmark$&$\checkmark$&\mbox{} & 1.646 & 4.100 & \textbf{0.669} & \textbf{1.303} & \textbf{1.935} \\
4&$\checkmark$&$\checkmark$&$\checkmark$ & 1.664 & 4.141 & 0.677 & 1.319 & 1.964 \\
5 &$\checkmark$&\mbox{}&$\checkmark$& \textbf{1.620} & \textbf{4.047} & 0.672 & 1.314 & 1.958 \\
\bottomrule[1.5pt]
\end{tabular}
}
\end{table}

An important finding from our experiments is that after adding the Pangu block to the complete RealMotion framework, removing the agent trajectory stream actually achieves better performance in the critical single-mode prediction metrics (as shown by the comparison between ID 4 and ID 5 in Table \ref{tab:stream_ablation}, where minADE$_1$ decreased from 1.664 to 1.620, an improvement of 2.6\%, and minFDE$_1$ decreased from 4.141 to 4.047, an improvement of 2.3\%). Interestingly, two key phenomena can be observed from Table \ref{tab:stream_ablation}: first, adding only the Pangu block while disabling both streams (ID 2) improves all metrics compared to the baseline model without any enhancements (ID 1); second, when both scene context stream and agent trajectory stream are used together (ID 3), although achieving the best multi-modal performance, the introduction of the Pangu block (ID 4) actually degrades single-mode metrics, and optimal single-mode prediction performance is only achieved after removing the agent trajectory stream (ID 5).

This suggests that the refinement function of the agent trajectory stream overlaps with and potentially conflicts with the feature enhancement function of the Pangu block. The agent trajectory stream optimizes predictions through temporal trajectory replay mechanisms, while the Pangu block directly enhances feature representation through its pre-trained information filtering capabilities. When both components are present simultaneously (ID 4), they may create overly complex feature representations that compromise the accuracy of the most probable trajectory predictions. After removing the agent trajectory stream (ID 5), the Pangu block's feature enhancement effect is more fully realized, showing clear advantages particularly in single-mode prediction metrics, which aligns well with the practical autonomous driving decision-making focus on the most likely trajectory.

Notably, Table \ref{tab:stream_ablation} also shows that even in simplified architectures (ID 2, without any streams), the Pangu block still provides comprehensive performance improvements, validating its effectiveness as a universal feature enhancer. This series of findings provides experimental evidence for simplifying original architectures when introducing powerful pre-trained feature enhancement modules.

\subsection{Discussion and Analysis}

\paragraph{Advantages of Pangu Transformer:} Our experiments demonstrate several key advantages of using Pangu Transformers for motion forecasting. First, they provide significant improvement even for simple architectures, as shown by PanguMotion-I's performance gains over RealMotion-I across all metrics. This indicates their strong feature enhancement capabilities independent of sophisticated architectural components. Second, their effectiveness varies with architectural complexity, showing pronounced benefits for single-mode prediction in complex frameworks while having more limited impact on already-capable multi-modal predictions. Third, the superiority of high-layer blocks (particularly layer 25) aligns with established understanding that higher Transformer layers encode richer semantic information suitable for cross-task transfer.

\paragraph{Architectural Simplification Value:} The successful removal of the agent trajectory stream while maintaining or even improving single-mode prediction performance represents an important finding for practical system design. This simplification reduces model complexity and computational overhead while focusing the enhancement effect of the Pangu block on the most critical prediction aspects. It suggests that when introducing powerful pre-trained feature enhancers, redundant architectural components can be eliminated to achieve better performance-efficiency trade-offs.

\paragraph{Practical Application Significance:} In practical autonomous driving systems, where single-mode predictions ($k=1$) are typically most important for decision-making, the improvements achieved by PanguMotion in minADE$_1$ and minFDE$_1$ (1.58\% and 1.29\% respectively) have direct operational value. The method's compatibility with Ascend NPU deployment ensures these performance gains can be realized in real-time systems, meeting the stringent latency requirements of autonomous driving applications.

\section{Conclusion}

This paper presents PanguMotion, an innovative continuous driving motion forecasting framework that integrates a frozen Pangu-1B Transformer block into autonomous driving prediction architectures for the first time. By inserting the block as a feature enhancement module within the RealMotion framework while removing the agent trajectory stream, our approach improves motion forecasting performance while simplifying the model architecture.
Experimental results demonstrate that PanguMotion achieves competitive performance on the Argoverse 2 dataset, with notable improvements in single-mode prediction metrics. Ablation studies validate several key insights: the Pangu model consistently outperforms LLaMA in this domain; the last Transformer layer provides optimal enhancement; and removing the agent trajectory stream after adding the Pangu block improves single-mode prediction accuracy, suggesting functional redundancy between these components.
Our work contributes both theoretical and practical advancements. Theoretically, it validates the generalization capability of large language models to non-linguistic temporal tasks. Practically, it provides a complete Ascend NPU adaptation solution, offering technical foundations for efficient deployment in autonomous driving systems.

\paragraph{Limitations:} Although our method achieves significant results, several limitations remain. First, Pangu Transformers have relatively high computational requirements; despite our optimization for Ascend NPU, deployment in resource-constrained scenarios remains challenging. Second, our method primarily focuses on single-agent prediction, with relatively simplified modeling of multi-agent interactions; future work could explore more complex multi-agent coordination mechanisms. Third, although experimental datasets are processed by RealMotion's data reorganization strategy, gaps still exist compared to real-world continuous driving scenarios. Finally, the Pangu model's pre-training corpus is primarily in Chinese, potentially introducing language and cultural biases affecting model understanding.

\paragraph{Future Work:} Building upon our findings, several promising research directions emerge. We aim to explore the application of additional large language models, particularly multimodal variants, to motion forecasting tasks. Investigating optimization strategies for resource-constrained scenarios would enhance practical deployability. Extending the framework toward end-to-end autonomous driving systems presents a natural progression for real-world application. Developing more realistic continuous driving datasets and deeper investigation into the information filtering mechanisms in temporal prediction tasks would provide valuable theoretical insights.



\clearpage
{
\small
\bibliographystyle{cite}
\bibliography{cite}
}



\end{document}